\documentclass[12pt]{spieman}  
\usepackage{amsmath,amsfonts,amssymb}
\usepackage{graphicx}
\usepackage{setspace}
\usepackage{tocloft}
\usepackage{import}
\usepackage{gensymb}

\title{Context-aware stacked convolutional neural networks for classification of breast carcinomas in whole-slide histopathology images}

\author[a,*]{Babak Ehteshami Bejnordi}
\author[b]{Guido Zuidhof}
\author[b]{Maschenka Balkenhol}
\author[b]{Meyke Hermsen}
\author[b]{Peter Bult}
\author[a]{Bram van Ginneken}
\author[a]{Nico Karssemeijer}
\author[a,b]{Geert Litjens}
\author[a,b]{Jeroen van der Laak}
\affil[a]{Radboud University Medical Center, Diagnostic Image Analysis Group, Department of Radiology and Nuclear Medicine, Nijmegen, The Netherlands}
\affil[b]{Radboud University Medical Center, Diagnostic Image Analysis Group, Department of Pathology, Nijmegen, The Netherlands}

\cftpagenumbersoff{figure}
\cftpagenumbersoff{table} 
\begin{document} 
\maketitle

\begin{abstract}
Automated classification of histopathological whole-slide images (WSI) of breast tissue requires analysis at very high resolutions with a large contextual area. In this paper, we present context-aware stacked convolutional neural networks (CNN) for classification of breast WSIs into normal/benign, ductal carcinoma in situ (DCIS), and invasive ductal carcinoma (IDC). We first train a CNN using high pixel resolution patches to capture cellular level information. The feature responses generated by this model are then fed as input to a second CNN, stacked on top of the first. Training of this stacked architecture with large input patches enables learning of fine-grained (cellular) details and global interdependence of tissue structures. Our system is trained and evaluated on a dataset containing 221 WSIs of H\&E stained breast tissue specimens. The system achieves an AUC of 0.962 for the binary classification of non-malignant and malignant slides and obtains a three class accuracy of 81.3\% for classification of WSIs into normal/benign, DCIS, and IDC, demonstrating its potentials for routine diagnostics.
\end{abstract}


{\noindent \footnotesize\textbf{*}Babak Ehteshami Bejnordi,  \linkable{Babak.EhteshamiBejnordi@Radboudumc.nl} }

\section{Introduction}
\label{sect:intro}  
Breast cancer is the most frequently diagnosed cancer among women worldwide. The most frequent subtype of breast cancer, invasive ductal carcinoma (IDC), accounts for more than 80\% of all breast carcinomas. IDC is considered to develop through sequential stages of epithelial proliferation starting from normal epithelium to invasive carcinoma via hyperplasia and ductal carcinoma in situ (DCIS) \cite{dupont1993breast}. DCIS is the pre-invasive stage of breast cancer in which the abnormal cells are confined to the lining of breast ducts. Accurate diagnosis of DCIS and IDC and their discrimination from benign diseases of the breast are pivotal to determine the optimal treatment plan. The diagnosis of these conditions largely depends on a careful examination of hematoxylin and eosin (H\&E) stained tissue sections under a microscope by a pathologist.

Microscopic examination of tissue sections is, however, tedious, time-consuming, and may suffer from subjectivity. In addition, due to extensive population-based mammographic screening for early detection of cancer, the amount of data to be assessed by pathologists is increasing. Computerized and computer-aided diagnostic systems can alleviate these shortcomings by assisting pathologists in diagnostic decision-making and improving their efficiency. Computational pathology systems can be used to sieve out obviously benign/normal slides and to facilitate diagnosis by pointing pathologists to regions highly suspicious for malignancy in whole slide images (WSI) as well as providing objective second opinions\cite{gurcan2009, Veta2014}. 

Numerous efforts have been undertaken to develop systems for automated detection of breast carcinomas in histopathology images\cite{Naik2008,doyle2008,dundar2011,Dong2014, Ehteshami16, Balazsi, cruz201, cruz2017accurate, Rezaeilouyeh, bejnordi2017deep}. Most of the existing algorithms for breast cancer detection and classification in histology images involve assessment of the morphology and arrangement of epithelial structures (e.g. nuclei, ducts). Naik et al. \cite{Naik2008} developed a method for automated detection and segmentation of nuclear and glandular structures for classification of breast cancer histopathology images. A large set of features describing the morphology of the glandular regions and spatial arrangement of nuclei was extracted for training a support vector machine classifier, yielding an overall accuracy of 80\% for classifying different breast cancer grades on a very small dataset containing a total of 21 pre-selected small regions of interest images. Doyle et al. \cite{doyle2008} further investigated the use of hand-crafted texture features for grading breast cancer histopathology images. Dundar et al.\cite{dundar2011} and Dong et al. \cite{Dong2014} developed automated classification systems based on an initial segmentation of nuclei and extraction of features to describe the morphology of nuclei or their spatial arrangement. While all of the previously mentioned algorithms were designed to classify manually selected regions of interest (mostly selected by expert pathologists), we \cite{Ehteshami16} proposed an algorithm based on multi-scale analysis of superpixels \cite{babakMulti} for automatic detection of ductal carcinoma in situ (DCIS) that operates at the whole slide level and distinguishes DCIS from a large set of benign disease conditions. Recently, Balazsi et al. \cite{Balazsi} proposed a system for detection of regions expressing IDC in WSIs. This system first divides the WSI into a set of homogeneous superpixels, and subsequently, uses a random forest classifier \cite{breiman2001} to determine if each region indicates cancer.

Recent advances in machine learning, in particular, deep learning \cite{Gu15,LeCu15}, have afforded state-of-the-art results in several domains such as speech \cite{hinton2012deep} and image recognition \cite{Kriz12, Russ14a}. Deep learning is beginning to meet the grand challenge of artificial intelligence by demonstrating human-level performance on tasks that require intelligence when carried out by humans \cite{silver2016mastering}. Obviating the need for domain-specific knowledge to design features, these systems learn hierarchical feature representations directly from data. On the forefront of methodologies for visual recognition tasks are convolutional neural networks (CNN). A CNN is a type of feed-forward neural network defined by a set of convolutional and fully connected layers. The emergence of deep learning, in particular, CNN, has also energized the medical imaging field \cite{litjens2017survey} and enabled development of diagnostic tools displaying remarkable accuracy \cite{gulshan2016development,esteva2017dermatologist,Camelyon16-2}, to the point of reaching human-level performance. These motivate the use of CNNs for detection and/or classification of breast cancer in breast histopathology images. 

Cruz et al. \cite{cruz201} proposed the first system using a CNN to detect regions of IDC in breast WSIs. In contrast to the modern networks that use very deep architectures to improve recognition accuracy, the utilized network was a 3-layer CNN.  Due to computational constraints, the model was only trained to operate on images downsampled by a factor of 16. In a recent publication\cite{cruz2017accurate}, the authors obtained comparable performance when training and validating their system on a multi-center cohort. Rezaeilouyeh et al. \cite{Rezaeilouyeh} trained a multi-stream 5-layer CNN taking as input a combination of RGB images, and magnitude and phase of shearlet coefficients. In all these works, the models were evaluated at the patch-level. In a recent work\cite{bejnordi2017deep}, we demonstrated the discriminating power of features extracted from tumor-associated stromal regions identified by a CNN for classifying breast WSI biopsies into invasive or benign.

\begin{figure}
\label{fig:lesions}
\begin{center}
\begin{tabular}{c}
\includegraphics[width=0.9\linewidth]{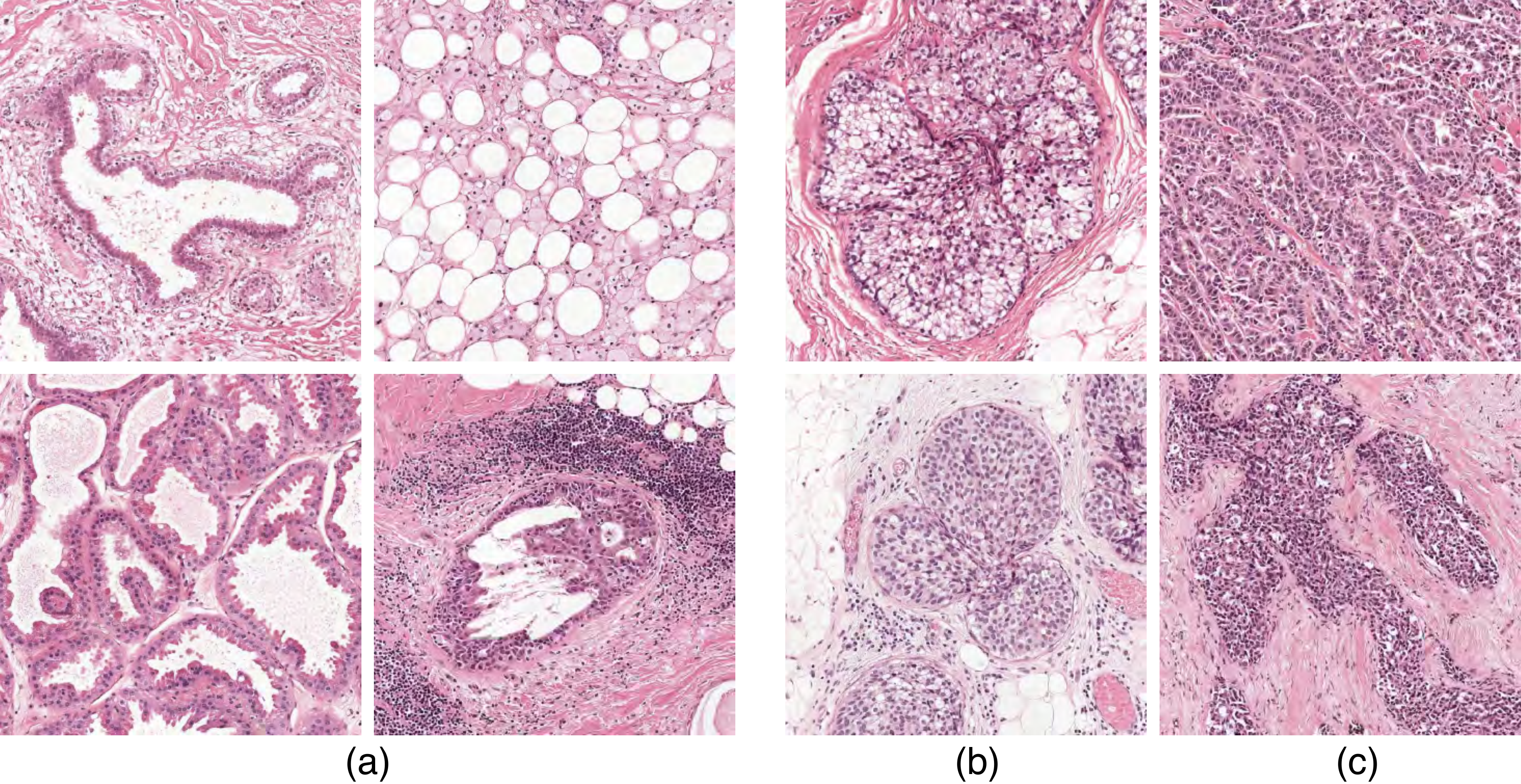}
\end{tabular}
\end{center}
\caption 
{Example of breast tissue strutures/lesion. (a) Normal tissue and benign lesions. (b) Ductal carcinoma in situ (DCIS). (c) Invasive ductal carcinoma (IDC).} 
\end{figure}

Different from the above mentioned approaches, the aim of the present study is to develop a system for WSI classification of breast histopathology images into three categories: normal/benign, DCIS and IDC categories. This problem is particularly difficult because of the wide range of appearances of benign lesions as well as the visual similarity of DCIS lesions to invasive lesions at the cellular level. Figure \ref{fig:lesions} shows some examples of lesions in our dataset. A system capable of discriminating these three classes, therefore, needs to use high-resolution information for discriminating benign lesions from cancer along with contextual information to discriminate DCIS from IDC. To develop a system that will work in a clinical setting, this study uses WSIs rather than manually extracted regions. Also, the cases in the `non-malignant' category contained many of the common benign lesions, as they appear in pathology practice.

To this end, we introduce context-aware convolutional neural networks for classification of breast histopathology images. First, we use a deep CNN which uses high pixel resolution information to classify the tissue into different classes. To incorporate more context to the classification framework, we feed a much larger patch to this model at test time. The feature responses generated by this model are then input to a second CNN, stacked on top of the first. This stacked network uses the compact, highly informative representations provided by the first model, which, together with the information from surrounding context, enables it to learn the global interdependence of various structures in different lesion categories. The performance of our system is evaluated on a large breast histopathology cohort comprising 221 WSIs from 122 patients.

\section{Methods}
\label{sect:methods}  
\subsection{Overview of the system}
The main challenge in the design of our classification framework is that the appearance of many benign diseases of the breast (e.g. usual ductal hyperplasia) mimic that of DCIS, hence requiring accurate texture analysis at the cellular level. Such analysis, however, is not sufficient for discrimination of DCIS from IDC. DCIS and IDC may appear identical on cellular examination but are different in their growth patterns which can only be captured through the inclusion of larger image patches containing more information about the global tissue architecture. Because of computational constraints, however, it is not feasible to train a deep CNN with large patches at high resolution that contain enough context.

Our method for classification of breast histopathology WSIs overcomes these problems through sequential analysis with a stack of CNNs. The key components of our classification framework, including the CNN used for classification of high-resolution patches, the stacked CNN for producing dense prediction maps, and a WSI labeling module are detailed in the following sections.

\subsection{Deep CNN for classification of small high-resolution patches}
\begin{figure}
\begin{center}
\begin{tabular}{c}
\includegraphics[width=0.99\linewidth]{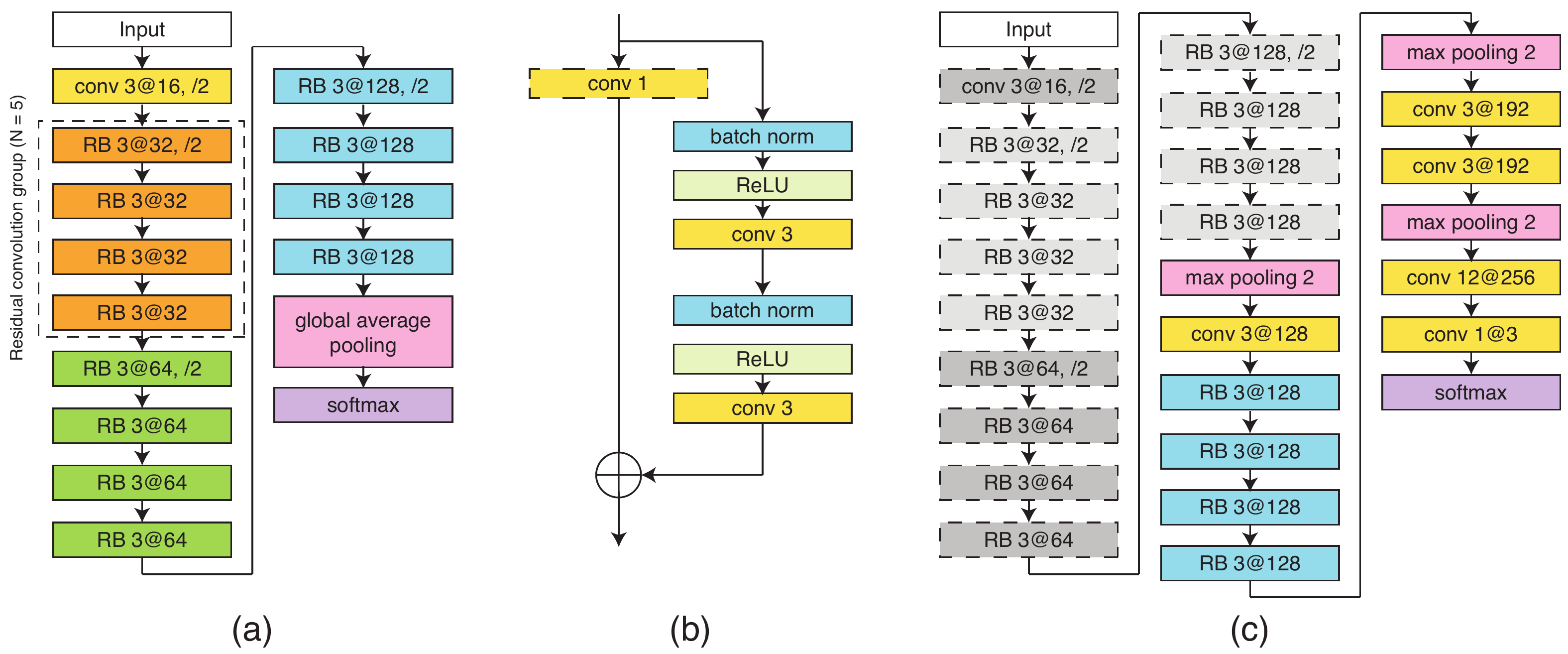}
\end{tabular}
\end{center}
\label{fig:Net}
\caption 
{ Architectures used for patch classification. (a) The \textit{WRN-4-2} architecture used for classification of $224\times224$ input patches. This architecture consists of an initial convolutional layer that is followed by three residual convolution groups (each of size N=4 residual blocks), followed by global average pooling and a softmax classifier. Downsampling is performed by the first convolutional layers in each group with a stride of 2 and the first convolutional layer of the entire network. Here, \textit{Conv 3@32} is a convolutional layer with a kernel size of $3\times 3$, and 32 filters. (b) The Residual Block (RB) used in this paper. Batch normalization and ReLU precede each convolution. $\oplus$ indicates an element-wise sum. Note that the $1\times1$ convolution layer is only used in the first convolutional layer of each Residual convolution group. (c) Architecture of the CAS-CNN with input size of $768\times768$. The weights of the components with dotted outlines are taken from the previously trained \textit{WRN-4-2} network, and are no longer updated during training.} 
\end{figure}

Inspired by the recent successes of deep residual networks \cite{he2016deep} for image classification, we trained and evaluated the performance of this CNN for classification of small high-resolution patches. We applied an adaptation of the ResNet architecture called wide ResNet as proposed by Zagoruyko et al. \cite{ZagoruykoK16}. This architecture has two hyperparameters: $N$ and $K$ determining the depth and width of the network. We empirically chose $N=4$ and $K=2$ to as a trade-off between model capacity, training speed and memory usage. Hereafter, we denote this network as \textit{WRN-4-2} (see figure \ref{fig:Net}). This network takes as input patches of size $224\times224$. Zero padding was used before each convolutional layer to keep the spatial dimension of feature maps constant after convolution.

The goal of this step was to transfer the highly informative feature representations learned by this network produced at its last convolutional layer to a stacked network which is described next. 

\subsection{Context-aware Stacked CNN (CAS-CNN)}
In order to increase the context available for dense prediction, we stack a second CNN on top of the last convolutional layer of the previously trained \textit{WRN-4-2} network. The architecture of the stacked network, as shown in figure \ref{fig:Net}, is a hybrid between the wide ResNet architecture and the VGG architecture \cite{simonyan2014very}. CAS-CNN is fully convolutional and enables fast dense prediction due to re-using of overlapping convolutions during inference. All the parameters of the \textit{WRN-4-2} network were fixed during training. Despite being trained with fixed input patches of size $224\times224$, because of being a fully convolutional network, \textit{WRN-4-2} can take a larger patch size during training of the stacked network, and consequently produce feature maps with larger spatial dimensions. Moreover, because of fixing the parameters of \textit{WRN-4-2}, the intermediate feature maps of this network do not need to be stored during backpropagation of the gradient. This allowed us to train stacked networks with much larger effective patch sizes. Consequently, we trained 3 networks with patch sizes of $512\times512$, $768\times768$, and $1024\times1024$.

Producing the dense prediction for a given WSI involved sliding the stacked network over the WSI with a stride of 224.

\subsection{WSI labeling}
Given a prediction map produced by the stacked network, we extracted a set of features describing global information about the lesions and their architectural distribution for subsequent classification into normal/benign, DCIS, or IDC. To this end, the probability map was transformed into a three label map, by assigning the class with the highest probability for every pixel. The three label map could contain several connected components for different object classes which were used for extracting features. Next, we describe the set of features extracted for WSI labeling.

\subsubsection{Global lesion features} These include the fraction of pixels classified as benign, DCIS, IDC, or cancerous (DCIS and IDC combined) with respect to all non-background pixels, along with the fraction of DCIS, and IDC labeled pixels with respect to all cancerous pixels. We additionally computed a convex-hull area feature for IDC detected lesions. IDC lesions usually appear as a large connected mass. As such, we constructed a convex hull of all IDC detected connected components in the WSI and computed the area ratio between the pixels labeled as IDC and the area of the convex hull. In case multiple tissue sections were present in the WSI, we took the average of these measures over different tissue sections. Note that IDC labeled connected components with an area smaller than 1500${\mu m}^2$ were discarded as false positives prior to computation of the convex hull feature. At the end, we computed the average area of DCIS connected components as well as IDC connected components as our two final global features. 

\subsubsection{Architectural features} These features describe the spatial distribution of DCIS and IDC lesions in the WSI. They were extracted from the area-Voronoi diagram \cite{bejnordi2013novel} and Delaunay triangulation (DT). We built these graphs for DCIS and IDC lesions independently. The seed points for constructing the graphs were the center of the connected components representing DCIS or IDC lesions. 

The set of features computed for each area-Voronoi region includes area, eccentricity, the area ratio of the Voronoi region and the total tissue area, and the area ratio of the lesion inside the Voronoi region and the Voronoi region itself. Per WSI, we computed the mean, median and standard deviation of these Voronoi area metrics. Additionally, we added the area of the largest Voronoi region to the feature set.

The features extracted for each of the nodes in the DT include the number of neighbors that are closer than a certain threshold to the node (threshold = 1500$\mu m$), and the average distance of these neighbors to the node. We computed the mean, median and standard deviation of these values as features. Additionally, we added the highest average node distance in the DT to the feature set.

Overall, a total of 57 features were extracted, which were used as input to two random forest classifiers \cite{breiman2001} with 512 decision trees: one for 3 class classification of WSIs and the other for binary classification of the WSIs into normal/benign versus cancerous (DCIS and IDC). We tuned the parameters of the classifiers by cross-validation on the combined set of train and validation WSIs.

\section{Experiments}
\label{sect:experiments}  
\subsection{Data}
We conducted our study on a large cohort comprising 221 digitized WSIs of H\&E stained breast tissue sections from 122 patients, taken from the pathology archive. Ethical approval was waived by the institutional review boards of the Radboud University Medical Center because all images were provided anonymously. All slides were stained in our laboratory and digitized using the 3DHISTECH Pannoramic 250 Flash II digital slide scanner with a 20X objective lens. Each image has square pixels of size $0.243\mu m \times 0.243\mu m$ in the microscope image plane.

Each slide was reviewed independently by a breast pathologist (PB) and assigned a pathological diagnosis. Overall, the dataset contains 100 normal/benign, 69 DCIS, and 55 IDC WSIs. Two human observers (MB and MH) annotated DCIS and IDC lesions using the Automated Slide Analysis Platform (ASAP) \cite{ASAP}. All the annotations were verified by the breast pathologist. Note that the slide labels were assigned according to the worst abnormality condition in the WSI. Therefore, a slide with the IDC label may contain both IDC and DCIS lesions.

We split this cohort into three separate sets: one for fitting classification models, one for intermediate validation and model selection, and one set for final evaluation of the system (test set). The training, validation, and test sets had 118 (50 normal/benign, 38 DCIS, and 30 IDC), 39 (19 normal/benign, 11 DCIS, and 9 IDC), and 64 (31 normal/benign, 20 DCIS, and 13 IDC) WSIs, respectively. There was no overlap at the slide- and patient-level between the three sets. The benign/normal category included 15 normal and 85 benign WSIs comprising fibroadenoma (14), ductal hyperplasia (11), adenosis (8), fibrosis (8), fibrocystic disease (8), duct ectasia (7), hamartoma (7), pseudo angiomatous stromal hyperplasia (5), sclerosing lobular hyperplasia (5), and mixed abnormalities (12). The WSIs from these 10 benign categories and the normal class were proportionally distributed in the training, validation, and test sets. Note that the relative occurrence of these lesions in our dataset is comparable to that encountered in routine diagnostics.

\subsection{Training protocols for CNNs}
We preprocessed all the data by scaling the pixel intensities between 0 and 1 for every RGB channel of the image patch and subtracting the mean RGB value that was computed on the training set. The training data was augmented with rotation, flipping, and jittering of the hue and saturation channels in the HSV color model.

Patches were generated on-the-fly to construct mini-batches during training and validation, by random selection of samples from points inside the contour of annotations for each class. For each mini-batch, the number of samples per class was determined with uniform probabilities.

Both WRN-4-2 and CAS-CNN were trained using Nesterov accelerated gradient descent. The weights of all trainable layers in the two networks were initialized using He initialization \cite{he2015}. Initial learning rates of 0.05 and 0.005 were used for WRN-4-2 and CAS-CNN, respectively. The learning rates were multiplied by 0.2 after no better validation accuracy was observed for a predefined number of consecutive epochs which we denote as epoch patience ($E_p$). The initial value for $E_p$ was set to 8 and increased by 20\% (rounded up) after every reduction in learning rate.  We used a mini-batch size of 22 for the WRN-4-2 and 18 for the CAS-CNN trained with patches of size $512\times512$ and $768\times768$. The network trained on $1024\times1024$ patches had a greater memory footprint and was trained with mini-batches of size 10.

Training of the WRN-4-2 involved one round of hard negative mining. Unlike the annotation of DCIS and IDC regions, the initial manual annotation of normal/benign areas was based on an arbitrary selection of visually interesting areas (e.g. areas that visually resembled cancer). These regions are not necessarily difficult for our network. In addition, some of the more difficult to classify benign regions could be underrepresented in our training set. We, therefore, enriched our training dataset by automatically adding all false positive regions in normal/benign training WSIs resulted by our initially trained WRN-4-2 model.

\subsection{Empirical evaluation and results}
We evaluated the performance of our system for classifying the WSIs into normal/benign, DCIS, and IDC categories using the accuracy measure and Cohen's kappa coefficient \cite{cohen1960coefficient}. We additionally measured the performance of our system for the binary classification of normal/benign versus cancer (DCIS and IDC combined) WSIs.  

As an intermediate evaluation, we began with measuring the performance of the WRN-4-2 for the binary and 3-class problems at the patch level (see in table \ref{PureNet_patchbased}). These are only results on the validation set, as this network is not used for producing the dense prediction maps individually. As can be seen, the model performs significantly better for the two class problem with an accuracy of 0.924 compared to the three class accuracy of 0.799 for the 3-class problem. This could be explained by the fact that WRN-4-2 only operates on small patches of size  $224\times224$ and does not have enough context for a more accurate discrimination of the three classes.

\begin{table}[h]
\renewcommand{\arraystretch}{1.1}
\caption{Patch-level accuracy for different networks on the validation set}
\label{PureNet_patchbased}
\centering
\begin{tabular}{|cccc|}
\hline
\textsc{Classification}&\textsc{Patch Size}&\textsc{Architecture}&\textsc{Accuracy}\\
\hline
\textit{Normal/Benign, Cancer}& $224\times224$&WRN-4-2& 0.9241\\
\hline
\textit{Normal/Benign, DCIS, IDC}&$224\times224$&WRN-4-2& 0.7995\\
\hline
\textit{Normal/Benign, DCIS, IDC}&$512\times512$&CAS-CNN& 0.8797\\
\hline
\textit{Normal/Benign, DCIS, IDC}&$768\times768$&CAS-CNN& 0.9050\\
\hline
\textit{Normal/Benign, DCIS, IDC}&$1024\times1024$&CAS-CNN& 0.9135\\
\hline
\end{tabular}
\end{table}

The results for the performance of the CAS-CNN on the validation set for the 3-class problem are shown in Table \ref{PureNet_patchbased}. The 3-class accuracy of this network was considerably improved compared to that of the WRN-4-2 at the patch level. We also observe that increasing the training patch-size leads to better performance. Accuracies of 0.872, 0.905, and 0.914 were obtained for the CAS-CNN networks trained on $512\times512$, $768\times768$, and $1024\times1024$ patches, respectively.

\begin{figure}
\begin{center}
\begin{tabular}{c}
\includegraphics[height=7cm]{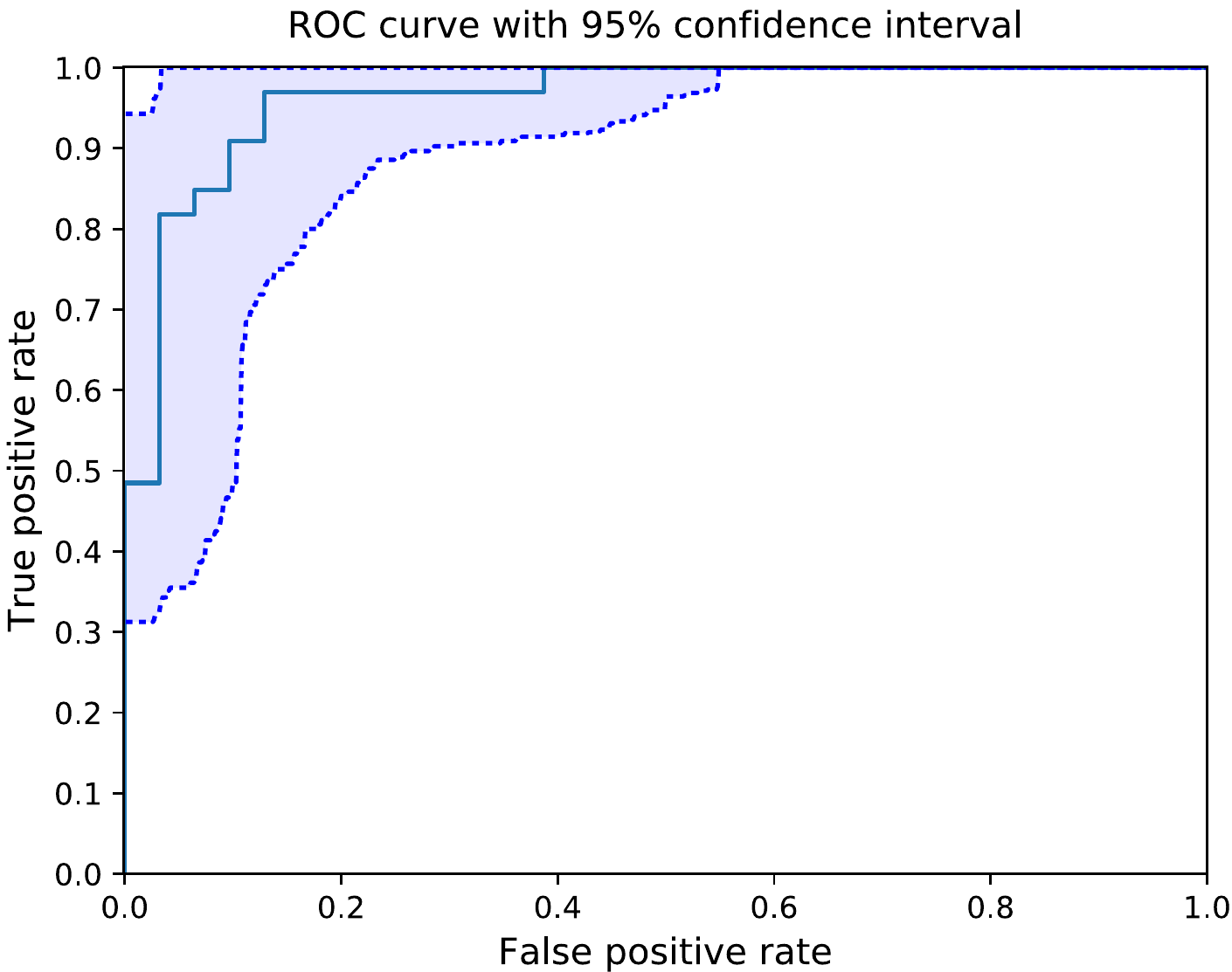}
\end{tabular}
\end{center}
\caption 
{ \label{fig:ROC}
ROC curve of the proposed system for binary classification of the WSIs in the test set into normal/benign and cancer (DCIS and IDC). } 
\end{figure}

Due to heavy computational costs of the network operating on $1024\times1024$ patches, the CAS-CNN network trained on $768\times768$ was ultimately selected for producing dense prediction maps. The results of the random forest classifier for WSI classification on the test set of our dataset are presented in table \ref{table_results_single_label}. For the binary classification task, our system achieves an AUC of 0.962. The accuracy and kappa values were 0.891 and 0.781, respectively. The ROC curve of the system for binary classification of WSIs into cancer versus normal/benign is shown in figure \ref{fig:ROC}.

\begin{table}[h]
\renewcommand{\arraystretch}{1.1}
\caption{Results of whole-slide image label prediction on the test set}
\label{table_results_single_label}
\centering
\begin{tabular}{|lrrr|}
\hline
\textsc{Labels}&\textsc{Acc}&\textsc{Kappa}&\textsc{AUC}\\
\hline
\textit{Benign, Cancer}&0.891&0.781&0.962 \\
\textit{Benign, DCIS, IDC}&0.813&0.700&- \\
\hline
\end{tabular}
\end{table}

\begin{figure}
\begin{center}
\begin{tabular}{c}
\includegraphics[width=0.9\linewidth]{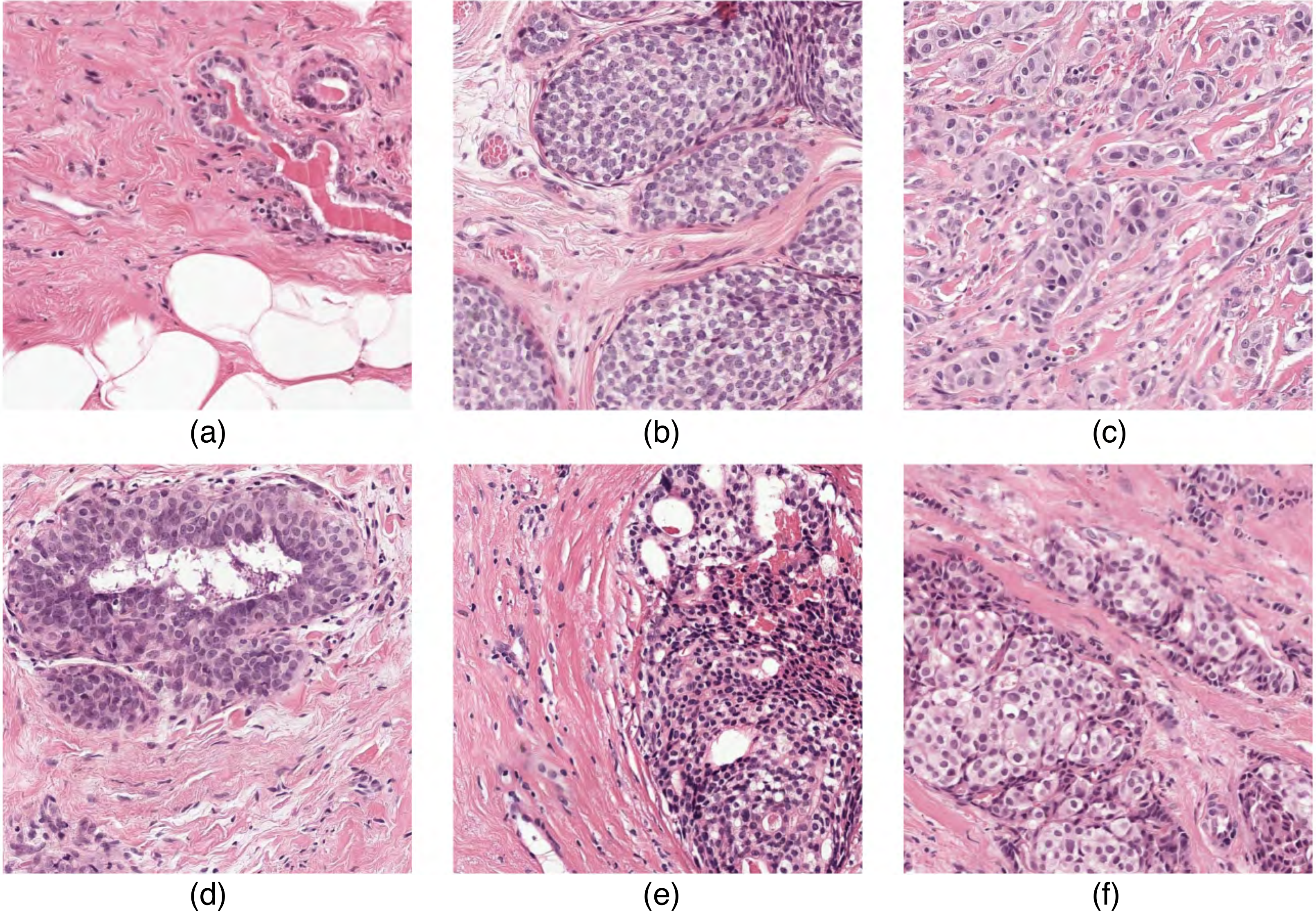}
\end{tabular}
\end{center}
\caption 
{ \label{fig:misclassifications}
Examples of correctly and incorrectly classified patches for different types of lesions. (a-c) correctly classified normal, DCIS, and IDC regions, respectively. (d) a benign lesion (usual ductal hyperplasia) misclassified as DCIS. (e) A DCIS lesion misclassified as normal/benign. (f) IDC misclassified as DCIS.} 
\end{figure}

The system achieves an overall accuracy and kappa value of 0.813 and 0.700 for three class classification of WSIs. The confusion matrix of the test set predictions is presented in table \ref{table_confusion_test}. Figure \ref{fig:misclassifications} presents several examples of correctly and incorrectly classified image patches for different lesion classes.

\begin{table}[h]
\renewcommand{\arraystretch}{1.21}
\caption{Confusion matrix of test set predictions}
\label{table_confusion_test}
\centering
\begin{tabular}{|l|rrr|}
\hline
\textsc{}&\textsc{Benign}&\textsc{DCIS}&\textsc{IDC}\\
\hline
\textsc{Benign}&29&2&0\\
\textsc{DCIS}&4& 12&4\\
\textsc{IDC}&0& 2&11\\
\hline
\end{tabular}
\end{table}

\section{Discussion and Conclusion}
\label{sect:diss}  
In this paper, we presented a context-aware stacked CNN (CAS-CNN) architecture to classify breast WSIs. To the best of our knowledge, this is the first approach investigating the use of deep CNNs for multi-class classification of breast WSIs into normal/benign, DCIS, and IDC categories.  CAS-CNN consists of two stages: in the first, we trained a CNN to learn cellular level features from small high-resolution patches and in the second, we stacked a fully convolutional network on top of this to allow for incorporation of global interdependence of structures to facilitate predictions in local regions. Our empirical evaluation demonstrates the efficacy of the proposed approach in incorporating more context to afford a high classification performance. CAS-CNN trained on large input patches outperforms the wide ResNet trained with input patches of size $224\times224$ by a large margin and consistently yields better results when trained with larger input patches.

Our system achieves an AUC of 0.962 for the binary classification of normal/benign slides from cancerous slides. This is remarkable, given the existence of 10 benign categories in the dataset, demonstrating the potential of our approach for pathology diagnostics. Based on the achieved performance on an independent test set, this system could be used to sieve out approximately 50\% of obviously normal/benign slides on our dataset without missing any cancerous slides.

The performance of the system on the 3-class classification of WSIs was also very promising. An accuracy of 0.812 and a Kappa value of 0.700 were achieved. While discrimination of normal/benign slides from IDC slides was without any misclassification, errors in discriminating between normal/benign slides and DCIS slides, as well as DCIS and IDC slides were common. We postulate that the reason for these misclassifications is primarily because of the difficulty in discrimination of several benign categories such as usual ductal hyperplasia from DCIS which is also a source of subjective interpretation among pathologists. This could, in turn, be alleviated by obtaining more training data for these specific benign classes. The second reason could be the requirement of even larger receptive fields to enable discrimination of DCIS from invasive cancer. As seen in Table \ref{PureNet_patchbased}, the performance of CAS-CNN consistently improved with increasing patch size. However, this comes with increased computation time both during training and inference. One way to redress the problem could be the inclusion of additional downsampled patches with larger receptive fields as input to a multi-scale network \cite{ghafoorian2017deep} or using alternative architectures such as U-net\cite{ronneberger2015u, MP12079}. The final reason behind these errors lies in the fact that discrimination of certain DCIS patterns from IDC, purely based on H\&E staining, can be complex. As such, pathologists may use additional staining such as myoepithelial markers to differentiate between DCIS and IDC lesions\cite{foschini2000differential}.

Although the current system learns to exhibit some hue and saturation invariance, specialized stain standardization techniques exist\cite{bejnordi2016stain, khan2014nonlinear, macenko2009method} and have been shown to greatly improve CAD system performance \cite{ciompi2017importance, wang2016deep} by reducing the stain variations \cite{bejnordi2014quantitative}. It is likely that standardizing the WSIs would also improve generalization of the performance of our network.

Although our primary aim was to facilitate pathology diagnostics by discriminating between different breast lesion categories, our system could serve as an important first step for the development of systems that aim at finding prognostic and predictive biomarkers within malignant lesions \cite{TUPAC16}. This will be one of our major directions for future work.

\subsection*{Disclosures}
The authors have no potential conflicts of interest.

\acknowledgments 
The authors wish to acknowledge the financial support by the European Union FP7 funded VPH-PRISM project under grant agreement n\degree601040.


\bibliography{report}   
\bibliographystyle{spiejour}   


\end{document}